\documentclass{article}
\usepackage{PRIMEarxiv}

\usepackage[utf8]{inputenc} 
\usepackage[T1]{fontenc}    
\usepackage{url}            
\usepackage{booktabs}       
\usepackage{amsfonts}       
\usepackage{nicefrac}       
\usepackage{microtype}      
\usepackage{lipsum}
\usepackage{fancyhdr}       
\usepackage{graphicx}       
\graphicspath{{media/}}     

\usepackage{amsmath,amsfonts,amscd,amssymb,bm}
\usepackage{graphicx}
\usepackage{epstopdf}
\usepackage{overpic}
\usepackage{cancel}
\usepackage{rotating}
\usepackage{url}
\usepackage{caption}
\usepackage{color}
\usepackage{rotating}
\usepackage{multirow}
\usepackage{wrapfig}
\usepackage{mathtools}
\usepackage{subeqnarray}
\usepackage{setspace}
\usepackage{palatino} 
\usepackage[numbers,sort&compress]{natbib}
\usepackage{algorithm}
\usepackage{algpseudocode}
\usepackage{mathtools}
\usepackage[most]{tcolorbox}
\usepackage{mathabx}
\usepackage{subcaption}
\usepackage{soul}
\usepackage{listings}
\usepackage{xcolor}
\usepackage{svg}

\definecolor{codegreen}{rgb}{0,0.6,0}
\definecolor{codegray}{rgb}{0.5,0.5,0.5}
\definecolor{codepurple}{rgb}{0.58,0,0.82}
\definecolor{backcolour}{rgb}{0.95,0.95,0.92}
\definecolor{burntorange}{rgb}{0.8, 0.33, 0.0}

\lstdefinestyle{mystyle}{
    backgroundcolor=\color{backcolour},   
    commentstyle=\color{codegreen},
    keywordstyle=\color{burntorange},
    numberstyle=\tiny\color{codegray},
    stringstyle=\color{codepurple},
    basicstyle=\ttfamily\footnotesize,
    breakatwhitespace=false,         
    breaklines=true,                 
    captionpos=b,                    
    keepspaces=true,                 
    numbers=left,                    
    numbersep=5pt,                  
    showspaces=false,                
    showstringspaces=false,
    showtabs=false,                  
    tabsize=2
}
\usepackage{hyperref}
\hypersetup{
    colorlinks=true,
    linkcolor=blue,
    filecolor=magenta,      
    urlcolor=blue,
    citecolor=black,
    }

\DeclareMathOperator*{\argmin}{arg\rm{}min}

\title{\textbf{Sparsifying Parametric Models with L$_0$ Regularization}}

\author{Nicolò Botteghi \\
	Department of Applied Mathematics\\
	University of Twente\\
	Enschede, Netherlands \\
	\texttt{n.botteghi@utwente.nl}
	\And Urban Fasel \\
	Department of Aeronautics\\
	Imperial College London \\
	London, United Kingdom \\
	\texttt{u.fasel@imperial.ac.uk}
}

\begin{document}
\maketitle



This document contains an educational introduction to the problem of sparsifying parametric models with L$_0$ regularization \cite{louizos2018learning}. We utilize this approach together with dictionary learning to learn sparse polynomial policies for deep reinforcement learning to control parametric partial differential equations \cite{botteghi2024parametric}. 
The document is organized as follows: in Section \ref{sec:NN_L0} we introduce the L$_0$ regularization that we use in our method introduced in \cite{botteghi2024parametric}. In Section \ref{sec:problem_settings}, we introduce the general problem setting for sparsifying parametric models. In Section \ref{sec:general_framework}, we discuss in more detail all the critical steps to derive the L$_0$ regularization, and in Section \ref{sec:L0_DRL}, we show different ways to use the L$_0$ regularization in deep reinforcement learning. The code and a tutorial are provided here: \url{https://github.com/nicob15/Sparsifying-Parametric-Models-with-L0}.

\vspace{24pt}

\section{Sparsifying Neural Network Layers with L$_0$ Regularization}\label{sec:NN_L0}

To sparsify the weight/coefficient matrix $\Xi$, the differentiable L$_0$ regularization method introduced in \cite{louizos2018learning} can be used. The method relaxes the discrete nature of L$_0$ to allow efficient and continuous optimization. 

Let $\bm{d}$ be a continuous random variable distributed according to a distribution $p(\bm{d}| \boldsymbol{\psi})$, where $\boldsymbol{\psi}$ indicates the parameters of $p(\bm{d}| \boldsymbol{\psi})$. Given a sample from $\bm{d} \sim p(\bm{d}|\boldsymbol{\psi})$, we can define the hard-sigmoid rectification:
\begin{equation}
    \bm{z} = \min(\bm{1}, \max(\bm{0}, \bm{d})).
    \label{eq:hard-sigmoid-rectification}
\end{equation}
Equation \eqref{eq:hard-sigmoid-rectification} allows $\bm{z}$, i.e., the learnable binary gate, to be exactly zero.
Additionally, we can still compute the probability of the gate being active, i.e., non-zero, by utilizing the cumulative distribution function $P$:
\begin{equation}
    p(\bm{z}\neq 0| \boldsymbol{\psi}) = 1 - P(\bm{d} \leq 0|\boldsymbol{\psi}).
\label{eq:prob_gate_active}
\end{equation}
We choose as candidate distribution a binary concrete \cite{maddison2016concrete, jang2016categorical}. Thus, the random variable $\bm{d}$ is distributed in $(0,1)$ with probability density $p(\bm{d}|\boldsymbol{\psi})$, cumulative density $P(\bm{d}|\boldsymbol{\psi})$, and learnable parameters $\boldsymbol{\psi}=[\log \boldsymbol{\alpha}, \beta]$, with $\log \boldsymbol{\alpha}$ the location and $\beta$ the temperature parameter. The distribution can be stretched to the interval $(\gamma, \zeta)$, where $\gamma < 0$ and $\zeta > 1$. Then, the hard-sigmoid on the samples analogously to Equation \eqref{eq:hard-sigmoid-rectification} can be applied:
\begin{equation}
\begin{split}
\bm{u} &\sim \mathcal{U}(\bm{0}, \bm{1})\, , \\ 
\bm{d} &= \sigma((\log\bm{u} - \log (1-\bm{u}) + \log \boldsymbol{\alpha})/ \beta) \, , \\
\bar{\bm{d}} &= \bm{d}(\zeta - \gamma) + \gamma \, , \\
\bm{z} &= \min(\bm{1}, \max(\bm{0}, \bar{\bm{d}}))\, , \\
\end{split}
\label{eq:binary_concrete}
\end{equation}
where $\sigma$ corresponds to the sigmoid activation function.
We can now optimize the parameters $\boldsymbol{\psi}$ of the distribution by minimizing the probability of the gate being active (see Equation \eqref{eq:prob_gate_active}). This optimization problem can be seen as stretching the distribution in $(0,1)$. Using Equation \eqref{eq:prob_gate_active} and the binary concrete distribution in Equation \eqref{eq:binary_concrete}, we can conveniently introduce the L$_0$ regularization loss as:
\begin{equation}
    L_0(\boldsymbol{\psi}) = \sum_{j=1}^{|\boldsymbol{\xi}|} (1-P_{\bar{d}_j}(0|\boldsymbol{\psi})) = \sum_{j=1}^{|\boldsymbol{\xi}|} \sigma(\log \alpha_j - \beta \log \frac{\gamma}{\zeta}),
\label{eq:L0_loss}
\end{equation}
where $\boldsymbol{\xi}$ are the parameters of the model we want to sparsify. At test time, we can estimate the sparse parameters $\boldsymbol{\xi}^0$ by:
\begin{equation}
    \begin{split}
        \bm{z} &= \min(\bm{1}, \max(\bm{0}, \sigma(\log \boldsymbol{\alpha})(\zeta - \gamma) + \gamma))\, , \\
        \boldsymbol{\xi}^0 &= \boldsymbol{\xi} \odot \bm{z}.
    \end{split}
\end{equation}

In the following two sections, we introduce the above idea and derive the concepts in more detail.  

\section{Problem Settings}\label{sec:problem_settings}
Suppose we have a dataset $\mathcal{D}$ of input and output pairs $\{(x_1, y_1), \cdots, (x_N, y_N)\}$. We consider a standard supervised learning setting, e.g., regression or classification, with an L$_0$ regularization to promote sparsity of the parameters $\boldsymbol{\xi}$ of a generic parametric model $h:\mathcal{X}\rightarrow\mathcal{Y}; x \mapsto h(x;\boldsymbol{\xi})$, e.g., a neural network. We can write the loss function for such a problem as:
\begin{equation}
\begin{split}
    \mathcal{L}(\boldsymbol{\xi}) &= \frac{1}{N} \sum_{i=1}^N F(h(x_i;\boldsymbol{\xi}), y_i) + \lambda ||\boldsymbol{\xi}||_0, \hspace{10pt} ||\boldsymbol{\xi}||_0 = \sum_{j=1}^{|\boldsymbol{\xi}|}\mathbb{I}[\xi_j \neq 0]\, , \\
    &= \mathcal{L}_E + \lambda\mathcal{L}_C \, ,\\
\end{split}
    \label{eq:l0_reg}
\end{equation}
where $F(\cdot)$ corresponds to a generic loss function, e.g., mean-squared error or cross-entropy, $|\boldsymbol{\xi}|$ is the dimensionality of the parameter vector $\boldsymbol{\xi}$, $\lambda$ is a weighting/scaling factor, and $\mathbb{I}$ is the indicator function. The first term of the loss function $\mathcal{L}_E$ corresponds to the \textit{error loss} that measures how well the model fits the training data, while $\mathcal{L}_C$ corresponds to the \textit{complexity loss} that measures the complexity (or sparsity) of the model. The optimal parameters $\boldsymbol{\xi}^*$ can be found as:
\begin{equation}
    \boldsymbol{\xi}^*=\argmin_{\boldsymbol{\xi}}\mathcal{L(\boldsymbol{\xi})}.
    \label{eq:optimal_params}
\end{equation}
The L$_0$ norm penalizes nonzero entries of the parameter vector and encourages sparsity in $\boldsymbol{\xi}^*$.

\begin{figure}[h!]
    \centering
    \includegraphics[width=0.8\textwidth]{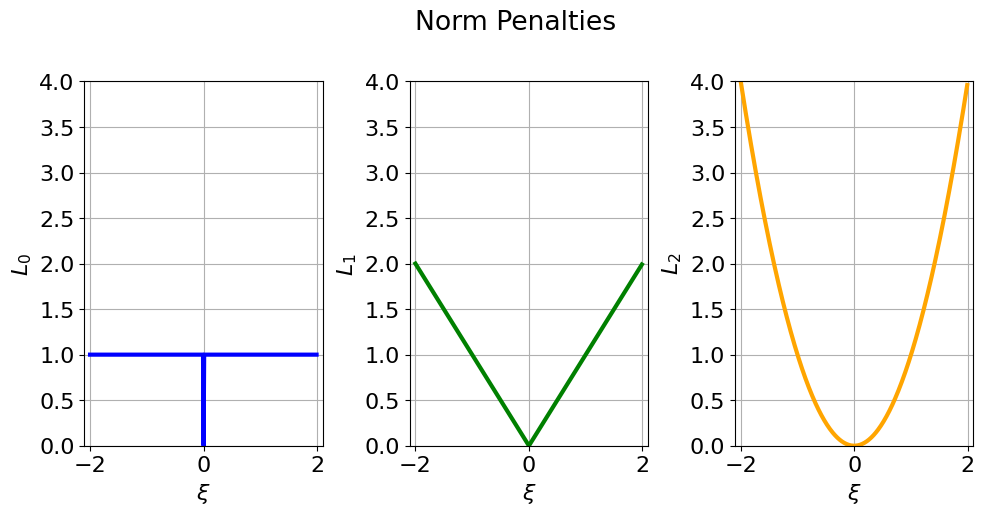}
    \caption{L$_0$, L$_1$, and L$_2$ norm penalties for a parameter $\xi$. Figure reproduced from \cite{loiseau2020data}.}
    \label{fig:different_norms}
\end{figure}

Unfortunately, the optimization constitutes an intractable brute force $2^{|\boldsymbol{\xi}|}$ combinatorial search due to the nondifferentiability of the L$_0$ complexity loss function. Usually, L$_1$ or Lasso, or L$_2$ norms are used as proxy of the L$_0$ norm as they are differentiable and can be used with gradient-based optimization techniques. Examples of these norms are visualized in Figure \ref{fig:different_norms}.

However,  L$_1$ and L$_2$ induce an undesirable shrinkage of the parameter values that is not introduced when using L$_0$.

\section{A General Framework for L$_0$ Regularization}\label{sec:general_framework}
Consider the L$_0$ norm under reparametrization of $\xi$:
\begin{equation}
    \xi^0_j = \xi_j z_j, \hspace{10pt} z_j \in \{0, 1\}, \hspace{10pt} \xi_j \neq 0, \hspace{10pt} ||\boldsymbol{\xi}||_0 = \sum_{j=1}^{|\boldsymbol{\xi}|}z_j\, ,
\end{equation}
where $z_j$ corresponds to the binary \textit{gate} that denotes whether a parameter is present or not. Under this reparametrization, the L$_0$ norm corresponds to the number of gates being active.
Using the Bernoulli distribution, we can reformulate the loss function in Equation \eqref{eq:l0_reg} as:
\begin{equation}
    \mathcal{L}(\boldsymbol{\xi}, \boldsymbol{\pi}) = \mathbb{E}_{q(\bm{z}|\boldsymbol{\pi})}\Big[\frac{1}{N} \sum_{i=1}^N F(h(x_i;\boldsymbol{\xi} \odot \bm{z}), y_i)\Big] + \lambda ||\boldsymbol{\xi}||_0, \hspace{10pt} ||\boldsymbol{\xi}||_0 = \sum_{j=1}^{|\boldsymbol{\xi}|}\pi_j\, ,
    \label{eq:l0_reg_bern}
\end{equation}
where $\odot$ indicated the element-wise product.
Analogously to Equation \eqref{eq:optimal_params}, we can find the optimal parameters $\boldsymbol{\xi}^*$ and $\boldsymbol{\pi}^*$ by solving:
\begin{equation}
    \boldsymbol{\xi}^*,\boldsymbol{\pi}^* =\argmin_{\boldsymbol{\xi}, \boldsymbol{\pi}}\mathcal{L(\boldsymbol{\xi}, \boldsymbol{\pi})}.
    \label{eq:optimal_params_bern}
\end{equation}
We could think of choosing a Bernoulli distribution over each gate $z_j$:
\begin{equation}
    q(z_j|\pi_j) = \text{Bern}(\pi_j)\, .
\end{equation}

\vspace{2pt}

\begin{tcolorbox}[enhanced jigsaw,breakable,pad at break*=1mm,
  colback=gray!5!white,colframe=gray!75!black,title=Bernoulli Distribution]
A Bernoulli distribution $\text{Bern}(\pi)$ describes the distribution of a random variable $X$ taking value 1 with probability $\pi$ or value 0 with probability $1-\pi$, with $\pi$ being the (learnable) parameter of the distribution:
\begin{equation}
    p(X=1) = 1-p(X=0) = \pi.
\end{equation}

\end{tcolorbox}

\vspace{2pt}

The optimization problem in Equation \eqref{eq:l0_reg_bern} and \eqref{eq:optimal_params_bern} is a special case of the variational lower bound over the parameters of the neural network involving spike and slab prior \cite{mitchell1988bayesian}. 

\subsection{Spike and Slab Distribution and Relation to Variational Inference}

The spike and slab distribution, shown in Figure \ref{fig:spikeandslab}, is considered the gold standard in sparsity-promoting Bayesian inference/linear regression. 

It is defined as a mixture of a delta spike at 0 and a continuous distribution over the real line, e.g., a standard Gaussian:
\begin{equation}
    p(z) = \text{Bern}(\pi), \hspace{10pt} p(\xi|z=0)=\delta(\xi), \hspace{10pt} p(\xi|z=1)=\mathcal{N}(\xi|0,1).
\end{equation}

\begin{figure}[h!]
    \centering
    \includegraphics[width=0.6\textwidth]{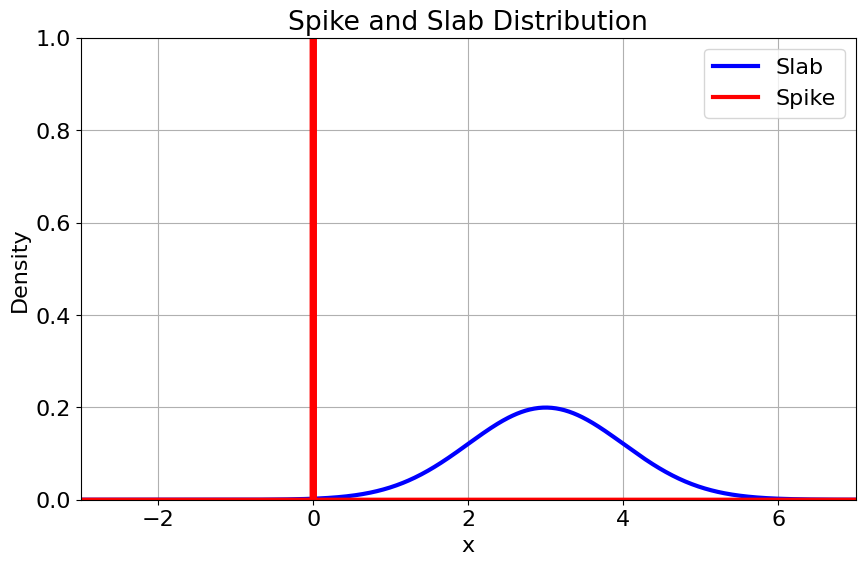}
    \caption{The spike and slab distribution.}
   \label{fig:spikeandslab}
\end{figure}
\label{fig:spikeandslab}

The true posterior distribution under the spike and slab prior is intractable. However, we can rely on variational inference \cite{kingma2013auto}.

Let $q(\xi, z)$ be a spike and slab approximate posterior over the parameters $\xi$ and the gate variables $z$. We can write the variational-free energy under the spike and slab prior and approximated posterior over a parameter vector $\boldsymbol{\xi}$ as:
\begin{equation}
\begin{split}
        \mathcal{F} &= -\mathbb{E}_{q(\bm{z})q(\boldsymbol{\xi}|\bm{z})}[\log p(\mathcal{D}|\boldsymbol{\xi})] + \sum_{j=1}^{|\boldsymbol{\xi}|} \text{KL}(q(z_j)||p(z_j)) \\
        &+ \sum_{j=1}^{|\boldsymbol{\xi}|}(q(z_j=1)\text{KL}(q(\xi_j | z_j=1)||p(\xi_j|z_j=1))) \\
        &+ \sum_{j=1}^{|\boldsymbol{\xi}|}(q(z_j=0)\text{KL}(q(\xi_j | z_j=0)||p(\xi_j|z_j=0)))\, , 
\end{split}
\label{eq:kl1}
\end{equation}
where $\text{KL}(q(\xi_j | z_j=0)||p(\xi_j|z_j=0))=0$ since the KL divergence of two Gaussian distributions with same mean and variance is equal to zero. Therefore, Equation \eqref{eq:kl1} can be rewritten as:
\begin{equation}
\begin{split}
        \mathcal{F} &= -\mathbb{E}_{q(\bm{z})q(\boldsymbol{\xi}|\bm{z})}[\log p(\mathcal{D}|\boldsymbol{\xi})] + \sum_{j=1}^{|\boldsymbol{\xi}|} \text{KL}(q(z_j)||p(z_j)) \\
        &+ \sum_{j=1}^{|\boldsymbol{\xi}|}(q(z_j=1)\text{KL}(q(\xi_j | z_j=1)||p(\xi_j|z_j=1)))\, ,
\end{split}
\label{eq:kl2}
\end{equation}
where the term $\text{KL}(q(z_j)||p(z_j))$ corresponds to the KL from the Bernoulli prior $p(z_j)$ and the Bernoulli approximate posterior $q(z_j)$, and $\text{KL}(q(\xi_j | z_j=1)||p(\xi_j|z_j=1))$ can be interpreted as the amount of information the parameter $\xi_j$ contains about the data $\mathcal{D}$, measured by the KL divergence from the prior $p(\xi_j|z_j=1)$. 
We can further simplify Equation \eqref{eq:kl2} by assuming, from an empirical Bayesian procedure, the existence of a hypothetical prior $p(\xi_j|z_j=1)$ for each parameter $\xi_j$ that adapts to $q(\xi_j|z_j=1)$ in a way that we need approximately $\lambda$ NATs (natural units of information) to transform $p(\xi_j|z_j=1)$ to that particular $q(\xi_j|z_j=1)$. Those $\lambda$ NATs are thus the amount of information that $q(\xi_j|z_j=1)$ can encode about the data if we had used $p(\xi_j|z_j=1)$ as the prior. This assumption translates into $\text{KL}(q(\xi_j | z_j=1)||p(\xi_j|z_j=1))=\lambda$. The coefficient $\lambda$ can be viewed as the amount of flexibility of that hypothetical prior. 
Eventually, if we consider optimizing over $\boldsymbol{\xi}$, instead of integrating, we can write the variational-free energy as:
\begin{equation}
        \mathcal{F} = -\mathbb{E}_{q(\bm{z})}[\log p(\mathcal{D}|\boldsymbol{\xi}\odot\bm{z})] + \sum_{j=1}^{|\boldsymbol{\xi}|} \text{KL}(q(z_j)||p(z_j)) + \lambda \sum_{j=1}^{|\boldsymbol{\xi}|}q(z_j=1)\, ,
\label{eq:kl3}
\end{equation}
where $\boldsymbol{\xi}$ corresponds to the optimized $\boldsymbol{\xi}$. Additionally, by using the positivity of the KL divergence, we can obtain the variational lower bound as:
\begin{equation}
        \mathcal{F} \geq -\mathbb{E}_{q(\bm{z})}[\log p(\mathcal{D}|\boldsymbol{\xi}\odot\bm{z})] + \lambda \sum_{j=1}^{|\boldsymbol{\xi}|}\pi_j\, ,
\label{eq:kl4}
\end{equation}
which is equivalent to Equation \eqref{eq:l0_reg_bern} if we take the negative log-probability of the data to be equal to the loss $\mathcal{L}(\cdot)$. This shows that the minimization of the L$_0$ norm is very close to the variational lower bound involving a spike and slab distribution over the parameters and a fixed cost/penalty for the parameters when the gates are active. Note that, if we are interested in quantifying uncertainties over the gate variables $\bm{z}$, we should optimize Equation \eqref{eq:kl3} (rather than \eqref{eq:kl4}) as this will properly penalize the entropy of $q(\bm{z})$. Equation \eqref{eq:kl3} also allows to incorporate prior information about the behaviour of the gates (e.g., being active $10\%$ of the time on average).

\subsection{Efficient Gradient-based Optimization of the L$_0$ Norm}
Minimizing Equation \eqref{eq:l0_reg_bern} is still not straightforward. While the second term of the loss is easy to minimize, the first term is still challenging due to the discrete nature of the gates $\bm{z}$, which does not allow for efficient gradient-based optimization. However, we can replace the Bernoulli distribution to smooth the objective function in Equation \eqref{eq:l0_reg_bern} and allow for efficient gradient-based optimization of the expected L$_0$ norm along with zeros in the parameters $\boldsymbol{\xi}.$ 

\begin{tcolorbox}[enhanced jigsaw,breakable,pad at break*=1mm,
  colback=gray!5!white,colframe=gray!75!black,title=Gradient Estimators]
In principle, it is possible to estimate the gradient using REINFORCE \cite{williams1992simple}. However, REINFORCE suffers from high variance of the estimates, requires auxiliary models, and multiple evaluations \cite{mnih2014neural, mnih2016variational, tucker2017rebar}. An alternative is to use the straight-through estimator \cite{bengio2013estimating} as done in \cite{srinivas2017training} or the concrete distribution \cite{gal2017concrete}. Unfortunately, the first method provides biased gradients (due to ignoring the Heaviside function in the likelihood during the gradient evaluation), while the second one does not allow for the gates to be exactly zero during the optimization (thus precluding the benefits of conditional computation \cite{bengio2013estimating}).
\end{tcolorbox}

\vspace{4pt}

Let $\bm{d}$ be a continuous random variable with a distribution $q(\bm{d})$ of parameters $\boldsymbol{\psi}$. We can now define the gates $\bm{z}$ as hard-sigmoid rectifications of $\bm{d}$ such that:
\begin{equation}
\begin{split}
    \bm{d} &\sim q(\bm{d}|\boldsymbol{\psi})\, , \\
    \bm{z} &= \min(\bm{1}, \max(\bm{0}, \bm{d}))) = g(\bm{d}).
\end{split}
\end{equation}
In this way, the gate is allowed to be exactly zero. Due to the underlying continuous random variable $\bm{d}$, we can still compute the probability of the gate being nonzero, i.e., active, from the cumulative distribution function (CDF) $Q(\bm{d}|\boldsymbol{\psi})$:
\begin{equation}
    q(\bm{z}\neq 0| \boldsymbol{\psi}) = 1 - Q(\bm{d}\leq 0|\boldsymbol{\psi})\, ,
    \label{eq:CDF}
\end{equation}
where $1 - Q(\bm{d}\leq 0|\boldsymbol{\psi})$ corresponds to the probability of $\bm{d}$ being positive. This allows to replace the binary Bernoulli gates in Equation \eqref{eq:l0_reg_bern} with the CDF in Equation \eqref{eq:CDF}:
\begin{equation}
    \mathcal{L}(\boldsymbol{\xi},\boldsymbol{\psi}) = \mathbb{E}_{q(\bm{d}|\boldsymbol{\psi})}\Big[\frac{1}{N}\sum_{i=1}^N F(h(x_i; \boldsymbol{\xi}\odot g(\bm{d})), y_i)\Big] + \lambda \sum_{j=1}^{|\boldsymbol{\xi}|}(1-Q(d_j\leq0|\psi_j)).
\label{eq:l0_reg_cdf}
\end{equation}
Analogously to Equation \eqref{eq:optimal_params_bern}, we can find the optimal parameters as:
\begin{equation}
   \boldsymbol{\xi}^*, \boldsymbol{\psi}^* = \argmin_{\boldsymbol{\xi}, \boldsymbol{\psi}}  \mathcal{L}(\boldsymbol{\xi},\boldsymbol{\psi})
\end{equation}

For a continuous distribution $q(\bm{d}|\boldsymbol{\psi})$ that allows for the reparametrization trick \cite{kingma2013auto}, we can express Equation \eqref{eq:l0_reg_cdf} as the expectation over a parameter-free noise distribution $p(\boldsymbol{\epsilon})$ and a deterministic and differentiable transformation $f(\cdot)$ of the parameters $\boldsymbol{\psi}$ and $\boldsymbol{\epsilon}$:
\begin{equation}
    \mathcal{L}(\boldsymbol{\xi},\boldsymbol{\psi}) = \mathbb{E}_{p(\boldsymbol{\epsilon})}\Big[\frac{1}{N}\sum_{i=1}^N F(h(x_i; \boldsymbol{\xi}\odot g(f(\boldsymbol{\psi}, \boldsymbol{\epsilon}))), y_i)\Big] + \lambda \sum_{j=1}^{|\boldsymbol{\xi}|}(1-Q(d_j\leq0|\psi_j))\, .
\label{eq:l0_reg_cdf2}
\end{equation}
This allows for a Monte Carlo approximation to the generally intractable expectation over the noise distribution $p(\boldsymbol{\epsilon})$:
\begin{equation}
\begin{split}
      \mathcal{L}(\boldsymbol{\xi},\boldsymbol{\psi}) &= \frac{1}{L}\sum_{l=1}^L \Big(\frac{1}{N}\sum_{i=1}^N F(h(x_i; \boldsymbol{\xi}\odot \bm{z}^{(l)}), y_i)\Big) + \lambda \sum_{j=1}^{|\boldsymbol{\xi}|}(1-Q(d_j\leq0|\psi_j))\, ,  \\\bm{z}^{(l)} &= g(f(\boldsymbol{\psi}, \boldsymbol{\epsilon}^{(l)}))\, , \hspace{10pt} \boldsymbol{\epsilon}^{(l)} \sim p(\boldsymbol{\epsilon}) . 
\end{split}
\label{eq:l0_reg_cdf3}
\end{equation}
Equation \eqref{eq:l0_reg_cdf3} is differentiable with respect to $\boldsymbol{\psi}$ and can be used with (stochastic) gradient-based optimization, while still allowing the parameters to be exactly zero. Additionally, we can choose an appropriate smoothing distribution $q(\bm{d}|\boldsymbol{\psi})$ for our problem. 
A choice that works well in practice is the \textit{binary concrete} \cite{maddison2016concrete, jang2016categorical}, with \textit{CONCRETE} derived from CONtinuous relaxation of disCRETE distribution \cite{maddison2016concrete}.

\subsubsection{The Hard-Concrete Distribution}
Assume to have  a binary concrete random variable $d$ distributed in $(0, 1)$ with a probability density function (PDF) $q(d|\psi)$:
\begin{equation}
    q(d|\psi) = \frac{\beta \alpha d^{-\beta-1}(1-d)^{-\beta-1}}{(\alpha d^{-\beta}+(1-d)^{-\beta})^2}\, ,
\end{equation}
 and a cumulative distribution function $Q(d|\psi)$:
 \begin{equation}
     Q(d|\psi) = \sigma\Big(\frac{\log d - \log(1-d)+\log \alpha}{\beta}\Big)\, ,
     \label{eq:cdf_binaryconcrete}
 \end{equation}
where the distribution has parameters $(\log \alpha, \beta)$, with $\log \alpha$ the location and $\beta$ the temperature, and $\sigma$ indicates the sigmoid activation function.

We can stretch the distribution in the $(\gamma, \xi)$ interval, with $\gamma < 0$ and $\xi > 1$ to obtain:
\begin{equation}
    \Bar{d}  = d(\xi - \gamma) + \gamma.
\end{equation}
The stretching of $d$ induces the following probability density function and cumulative distribution function $q(\Bar{d}|\psi)$ and $Q(\Bar{d}|\psi)$:
\begin{equation}
    \begin{split}
        q(\bar{d}|\psi) &= \frac{1}{|\xi - \gamma|}q\Big(\frac{\Bar{d}-\gamma}{\xi - \gamma}|\psi\Big)\, ,\\
        Q(\bar{d}|\psi) &= Q\Big(\frac{\Bar{d}-\gamma}{\xi - \gamma}|\psi\Big).\\
    \end{split}
\end{equation}

We can further rectify $\Bar{d}$ with the hard-sigmoid  such that:
\begin{equation}
    z = \min(1, \max(0, \Bar{d})).
\end{equation}
We obtain the following distribution over $z$, which is referred to as the \textit{hard-concrete distribution}:
\begin{equation}
    q(z|\psi) = Q_{\Bar{d}}(0|\psi) \delta(z) + (1- Q_{\Bar{d}}(1|\psi))\delta(z-1) + (Q_{\Bar{d}}(1|\psi)) - Q_{\Bar{d}}(0|\psi)))q_{\Bar{d}}(z|\Bar{d}\in (0,1), \psi)\, .
\end{equation}
This distribution is composed of a delta peak at zero with probability $Q_{\Bar{d}}(0|\psi))$, a delta peak at one with probability $1-Q_{\Bar{d}}(1|\psi))$, and a truncated version of $q_{\Bar{d}}(\Bar{d}|\psi)$ in the range $(0, 1)$.

\begin{figure}[h!]
    \centering
    \includegraphics[width=0.8\textwidth]{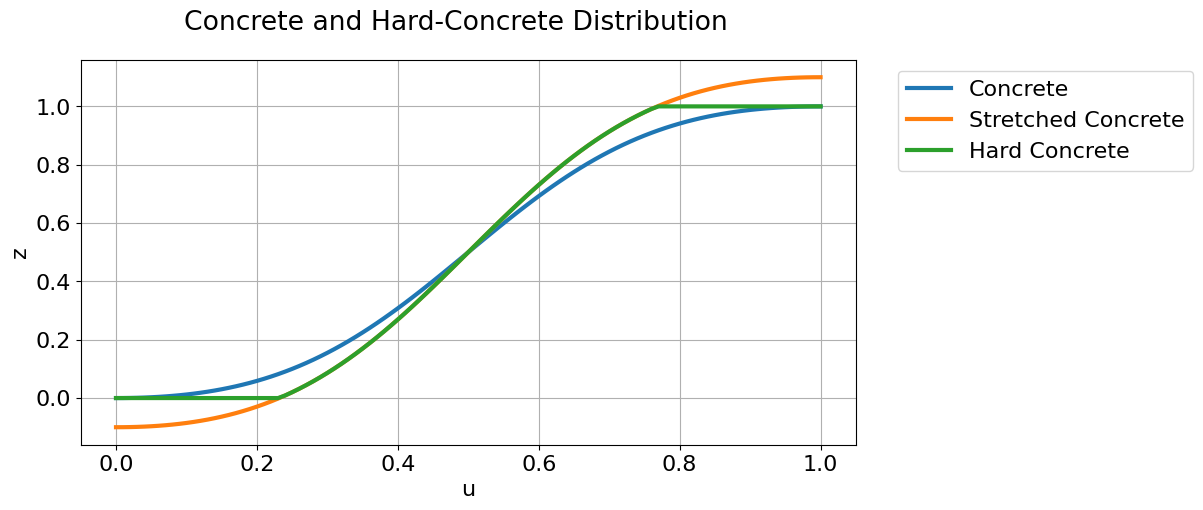}
    \caption{Gate $z$ for the binary concrete and hard-concrete distributions. Figure recreated from \cite{louizos2018learning}.}
    \label{fig:gating}
\end{figure}

In practice, we can sample a gate $z$ from the hard-concrete distribution by first applying the reparametrization trick, then sampling from a uniform distribution $\mathcal{U}(0,1)$, feeding the sample to the binary concrete CDF, stretching it, and finally passing it to the hard-sigmoid rectification:
\begin{equation}
    \begin{split}
        u &\sim \mathcal{U}(0, 1) \\ 
        d &= \sigma\Big(\frac{\log u - \log(1-u)+\log \alpha}{\beta}\Big) \\
        \Bar{d}  &= d(\xi - \gamma) + \gamma \\
        z &= \min(1, \max(0, \Bar{d})).
    \end{split}
\end{equation}
An example of a resulting gate $z$ is shown in Figure \ref{fig:gating}.

We can derive the CDF in Equation \eqref{eq:cdf_binaryconcrete} as:
\begin{equation}
    \begin{split}
        Q(d|\psi) &= \int_{-\infty}^s q(x|\psi) dx \, , \\
        &= \int_{-\infty}^s \frac{\beta \alpha x^{-\beta-1}(1-x)^{-\beta-1}}{(\alpha x^{-\beta}+(1-x)^{-\beta})^2} dx.\\
    \end{split}
\end{equation}
Solving the integral is very challenging due to the form of the PDF. However, we can rely on the Gumbel-Max trick \cite{maddison2016concrete}. We start from the observation that Bernoulli random variables are a special case of discrete distributions with states in $\{0, 1\}$. Consider a discrete distribution $D \sim \text{Discrete}(\alpha)$, with $\alpha \in (0, \infty)$ the parameter of the discrete distribution, and a two-state discrete random variable on $\{0, 1\}^2$ such that $D_1 + D_2 = 1$:
\begin{equation}
    Q(D_1=1) = \frac{\alpha_1}{\alpha_1 + \alpha_2}.
\label{eq:gumbel-max}
\end{equation}

\vspace{2pt}

\begin{tcolorbox}[enhanced jigsaw,breakable,pad at break*=1mm,
  colback=gray!5!white,colframe=gray!75!black,title=The Gumbel-Max Trick]
The Gumbel-Max trick proceeds as follows:
\begin{itemize}
    \item sample $U_k  \sim \mathcal{U}(0, 1)$ with $k=1, 2$,
    \item find the index $k$ maximizing $\log(\log U_k) + \log \alpha_k$, and
    \item set $D_k=1$ and the remaining $D_{i\neq k}=0$.
\end{itemize}
\end{tcolorbox}

\vspace{8pt}

Therefore, if we apply the Gumbel-Max trick, the case of $D_1=1$ corresponds to the event:
\begin{equation}
\begin{split}
    \log(\log U_1) + \log \alpha_1 &> \log(\log U_2) + \log \alpha_2 \, , \\
    G_1+ \log \alpha_1 &> \log(\log U_2) + \log \alpha_2 .
\end{split}
\end{equation}
The difference between $\log(\log U_1)=G_1$ and $\log(\log U_2)=G_2$ is a logistic distribution $L$ with CDF the logistic function $\sigma(x) = 1/(1+\exp^{-x})$:
\begin{equation}
    G_1 - G_2 \sim L(U) = \log U - \log(1-U) \, ,
\end{equation}
where $U \sim \mathcal{U}(0,1)$.
Therefore, we can rewrite Equation \eqref{eq:gumbel-max} as:
\begin{equation}
\begin{split}
        Q(D_1=1) &= Q(\log(\log U_1) + \log \alpha_1 > \log(\log U_2) + \log \alpha_2) \\
        &= Q(G_1 - G_2 + \log \alpha_1 - \log \alpha_2) \\
        &= Q( \log U - \log(1-U) + \log \alpha > 0)\, ,
\end{split}
\label{eq:gumbel_2}
\end{equation}
where $\log \alpha = \log \alpha_1 - \log \alpha_2$.

Eventually, we can define the binary concrete random variable as:
\begin{equation}
   Z = \sigma\Big(\frac{\log U - \log(1-U) + \log \alpha}{\beta}\Big)
\end{equation}
and its CDF as:
\begin{equation}
\begin{split}
       Q(z) &= Q(Z \leq z) = Q\Big(\sigma\Big(\frac{\log U - \log(1-U) + \log \alpha}{\beta}\Big)\leq z\Big) \\
       &= Q\Big(\frac{\log U - \log(1-U) + \log \alpha}{\beta}\leq \sigma(z)^{-1}\Big) \\
       &= Q\Big(\frac{\log U - \log(1-U) + \log \alpha}{\beta}\leq \log \Big(\frac{z}{1-z}\Big)\Big) \\
       &= Q\Big(\log U - \log(1-U) + \log \alpha \leq \beta\log \Big(\frac{z}{1-z}\Big)\Big) 
\end{split}
\label{eq:CDF_proof}
\end{equation}
where $\sigma(z)^{-1}=\log \Big(\frac{z}{1-z}\Big)$ is the inverse of $\sigma(z)$. Equation \eqref{eq:CDF_proof} is equivalent to \eqref{eq:gumbel_2}.

\newpage

\section{Deep Reinforcement Learning with L$_0$ Regularization}\label{sec:L0_DRL}
To conclude the tutorial, we show how to use the differentiable L$_0$ regularization in the context of deep reinforcement learning to learn transition models, reward models, and control policies. We use as test case the pendulum environment from OpenAI Gym \cite{openaigym} \url{https://www.gymlibrary.dev/environments/classic_control/pendulum/}.

\subsection{Learning Transition and Reward Models}

\subsubsection{Dataset Collection}

We start by creating a training set of 1000 episodes and a test set of 100 episodes. These dataset are obtained by applying a random policy to the environment for 200 steps per episodes. The code is shown and explained in Code \ref{lst:train_set} and \ref{lst:test_set}.

\vspace{8pt}

\begin{lstlisting}[language=Python, caption=Creation of training set., label={lst:train_set}]
import gymnasium as gym
from utils.replay_buffer import ReplayBuffer

# initialize gym environment
env = gym.make('Pendulum-v1', g=9.81)

# set maximum number of episodes and steps
max_episodes = 1000
max_steps = 200

# get dimension of the observations, actions, and memory buffer
obs_dim = env.observation_space.shape[0]
act_dim = env.action_space.shape[0]
buf_dim = int(max_episodes*max_steps)

# initialize training buffer (we will store the training data here)
training_buffer = ReplayBuffer(obs_dim=obs_dim, act_dim=act_dim, size=buf_dim)

# create training set
for episode in range(max_episodes):

    # reset the environment at the beginning of each episode
    observation, _ = env.reset() 
    
    for steps in range(max_steps+1):

        # select action according to a random policy
        action = env.action_space.sample() 

        # apply the action to the environment
        next_observation, reward, terminated, truncated, _ = env.step(action) 

        done = terminated or truncated

        # store data tuple in the training buffer
        training_buffer.store(observation, action, reward, next_observation, done)

        # set next observation as the current observation
        observation = next_observation

        if done:
            break

\end{lstlisting}

\begin{lstlisting}[language=Python, caption=Creation of test set., label={lst:test_set}]
# set maximum number of episodes 
max_episodes_test = 100

# set dimension of memory buffer
buf_dim = int(max_episodes*max_steps)

# initialize testing buffer (we will store the test data here)
testing_buffer = ReplayBuffer(obs_dim=obs_dim, act_dim=act_dim, size=buf_dim)

# create test set
for episode in range(max_episodes_test):

    # reset the environment at the beginning of each episode
    observation, _ = env.reset()
    
    for steps in range(max_steps + 1):

        # select action according to a random policy
        action = env.action_space.sample()

        # apply the action to the environment
        next_observation, reward, terminated, truncated, _ = env.step(action)

        done = terminated or truncated

        # store data tuple in the testing buffer
        testing_buffer.store(observation, action, reward, next_observation, done)

        # set next observation as the current observation
        observation = next_observation

        if done:
            break

\end{lstlisting}

\subsubsection{Transition and reward models}
We first show how to learn:
\begin{itemize}
    \item a transition model $T: \mathcal{S} \times \mathcal{A} \rightarrow \mathcal{S}$, mapping a state-action pair $(\bm{s}_t, \bm{a}_t)$ to the next state $s_{t+1}$, and
    \item a reward model $R: \mathcal{S} \times \mathcal{A} \rightarrow \mathbb{R}$, mapping a state-action pair $(\bm{s}_t, \bm{a}_t)$ to the reward $r_t$.
\end{itemize} 
We approximate $T$ and $R$ in three different ways:
\begin{itemize}
    \item using a fully-connected neural network with parameters $\boldsymbol{\xi}_{T}$ and $\boldsymbol{\xi}_{R}$, respectively,
    \item using a fully-connected neural network trained with L$_0$ regularization with parameters $\boldsymbol{\xi}_T$ and $\boldsymbol{\psi}_T$ and $\boldsymbol{\xi}_R$ and $\boldsymbol{\psi}_R$, and
    \item using a SINDy-like model \cite{brunton2016discovering} trained with L$_0$ regularization with parameters $\boldsymbol{\xi}_T$ and $\boldsymbol{\psi}_T$ and $\boldsymbol{\xi}_R$ and $\boldsymbol{\psi}_R$ (this is analogous to the sparse policy introduced in \cite{botteghi2024parametric}). We utilize a polynomial-feature library, a Fourier library, and a generalized library composed of polynomial and Fourier features.
\end{itemize}

The fully-connected neural network model is shown in Code \ref{lst:FCNN_model}. The models is composed of three layers with ELU activation for the first two. For learning transition and reward model, we utilize the same architecture.
\begin{lstlisting}[language=Python, caption=Fully-connected neural network model., label={lst:FCNN_model}]
class FCNN(nn.Module):
    def __init__(self, input_dim=3, output_dim=1, h_dim=256, use_bias=True):
        super(FCNN, self).__init__()

        self.use_bias = use_bias
        
        self.fc = nn.Linear(input_dim, h_dim, bias=use_bias)
        
        self.fc1 = nn.Linear(h_dim, h_dim, bias=use_bias)
        
        self.fc2 = nn.Linear(h_dim, output_dim, bias=use_bias)

    def forward(self, obs, act):
        # concatenate observation and action before feeding them to the input layer
        x = torch.cat([obs, act], dim=1)
        x = F.elu(self.fc(x))
        x = F.elu(self.fc1(x))
        # the output is either the next observation or the reward
        out = self.fc2(x)
        return out
\end{lstlisting}
Similarly to the fully-connected neural network, the sparse fully-connected neural network is composed of three layers with ELU activations. However, the layers include the L$_0$ mask introduced in \cite{louizos2018learning}.
\begin{lstlisting}[language=Python, caption=Sparse fully-connected neural network model., label={lst:sparseFCNN_model}]
class SparseFCNN(nn.Module):
    def __init__(self, input_dim=3, output_dim=1, h_dim=256, weight_decay=0.,                           droprate_init=0.5, temperature=2./3., lambda_coeff=1.):
        
        super(SparseFCNN, self).__init__()

        self.fc = L0Dense(in_features=input_dim, out_features=h_dim, bias=True,                             weight_decay=weight_decay, droprate_init=droprate_init,                            temperature=temperature, lamba=lambda_coeff, local_rep=False)
        
        self.fc1 = L0Dense(in_features=h_dim, out_features=h_dim, bias=True,                                weight_decay=weight_decay, droprate_init=droprate_init,                            temperature=temperature, lamba=lambda_coeff, local_rep=False)
        
        self.fc2 = L0Dense(in_features=h_dim, out_features=output_dim, bias=True,                           weight_decay=weight_decay, droprate_init=droprate_init,                            temperature=temperature, lamba=lambda_coeff, local_rep=False)

    def forward(self, obs, act):
        # concatenate observation and action before feeding them to the input layer
        x = torch.cat([obs, act], dim=1)
        x = F.elu(self.fc(x))
        x = F.elu(self.fc1(x))
        # the output is either the next observation or the reward
        out = self.fc2(x)
        return out
\end{lstlisting}
Eventually, in Code \ref{lst:L0SINDy_model}, we show the structure of the L$_0$ SINDy like model with three different feature libraries. The model makes use of the library features of the PySINDy library \url{https://github.com/dynamicslab/pysindy} and of a single sparse layer to learn the coefficients of each feature. The sparse layer allows for utilizing the L$_0$ regularization introduced in \cite{louizos2018learning} to learn a sparse linear combination of the nonlinear library features.
\begin{lstlisting}[language=Python, caption=L$_0$ SINDy-like model., label={lst:L0SINDy_model}]
    class L0SINDy_model(nn.Module):
    def __init__(self, input_dim=3, output_dim=1, weight_decay=0., droprate_init=0.5,                   temperature=2. / 3., lambda_coeff=1., degree=3, frequency=1,                           lib_type='polynomial'):
        
        super(L0SINDy_reward, self).__init__()

        if lib_type == 'polynomial':
            self.lib = PolynomialLibrary(degree=degree, include_bias=True,                                                 include_interaction=True)
            x = np.ones((1, input_dim))
            self.lib.fit(x)
            xf = self.lib.transform(x)
            coef_dim = xf.shape[1]

        if lib_type == 'fourier':
            self.lib = FourierLibrary(n_frequencies=frequency, include_sin=True,                                       include_cos=True, interaction_terms=True)
            x = np.ones((1, input_dim))
            self.lib.fit(x)
            xf = self.lib.transform(x)
            coef_dim = xf.shape[1]

        if lib_type == "polyfourier":
            poly_lib = PolynomialLibrary(degree=degree, include_bias=True,                                                 include_interaction=True)
            fourier_lib = FourierLibrary(n_frequencies=frequency, include_sin=True,                                        include_cos=True, interaction_terms=True)
            self.lib = GeneralizedLibrary([poly_lib, fourier_lib])
            x = np.ones((1, input_dim))
            self.lib.fit(x)
            xf = self.lib.transform(x)
            coef_dim = xf.shape[1]

        # single sparse layer learning the coefficents of each dictionary feature
        self.fc = L0Dense(in_features=coef_dim, out_features=output_dim, bias=False,                         weight_decay=weight_decay, droprate_init=droprate_init,                            temperature=temperature, lamba=lambda_coeff, local_rep=False)

    def forward(self, obs, act):
        # concatenate observation and action before feeding them to the input layer
        x = torch.cat([obs, act], dim=1)
        # compute the features of x using the chosen library functions
        xf = torch.from_numpy(self.lib.transform((x).cpu().numpy())).to(self.device)
        # feed the features to a single layer of sparse fully-connected neural network
        out = self.fc(xf)
        return out
\end{lstlisting}

\subsubsection{Training the Models}
We indicate with $\boldsymbol{\xi}$ the parameters of the different models and with $\boldsymbol{\psi}$ the L$_0$-mask parameters.
We train the models to minimize the mean-squared error between the predictions and the ground truth. In particular, we optimize the fully-connected neural network parameters with:
\begin{equation}
    \mathcal{L}(\boldsymbol{\xi}_T) = \mathbb{E}_{\bm{s}_t, \bm{a}_t, \bm{s}_{t+1}}[||\bm{s}_{t+1} - \hat{\bm{s}}_{t+1}||^2_2]\, ,
\end{equation}
and
\begin{equation}
    \mathcal{L}(\boldsymbol{\xi}_R) = \mathbb{E}_{\bm{s}_t, \bm{a}_t, r_t}[||r_t - \hat{r}_t||^2]\, ,
\end{equation}
where $\hat{\bm{s}}_{t+1}=T(\bm{s}_t, \bm{a}_t, \boldsymbol{\xi})$ is the prediction of the transition model $T$ and  $\hat{r}_t=R(\bm{s}_t, \bm{a}_t, \boldsymbol{\xi})$ is the prediction of the reward model $R$, respectively.

The models using the L$_0$ regularization are instead optimized according to Equation \eqref{eq:l0_reg_cdf3}:
\begin{equation}
    \mathcal{L}(\boldsymbol{\xi}_T, \boldsymbol{\psi}_T) = \mathbb{E}_{\bm{s}_t, \bm{a}_t, \bm{s}_{t+1}}[||\bm{s}_{t+1} - \hat{\bm{s}}_{t+1}||^2_2] + \lambda \text{L}_0(\boldsymbol{\psi}_T)\, ,
\end{equation}
and
\begin{equation}
    \mathcal{L}(\boldsymbol{\xi}_R, \boldsymbol{\psi}_R) = \mathbb{E}_{\bm{s}_t, \bm{a}_t, r_t}[||r_t - \hat{r}_t||^2] + \lambda \text{L}_0(\boldsymbol{\psi}_R).
\end{equation}

Code \ref{lst:dyn_model_training} and \ref{lst:rew_model_training}, we show the training loop for the transition and reward models.
\begin{lstlisting}[language=Python, caption=Transition model training loop., label={lst:dyn_model_training}]
def train_dynamics_model(model, optimizer, train_loader, batch_size,                                                num_training_iterations, l0=False):

    for _ in range(num_training_iterations):

        # random sample a batch of data
        data = train_loader.sample_batch(batch_size)

        obs = torch.from_numpy(data['obs']).cuda()
        next_obs = torch.from_numpy(data['next_obs']).cuda()
        act = torch.from_numpy(data['act']).cuda()

        optimizer.zero_grad()

        # predict the next observation using the parametric model
        pred_next_obs = model(obs, act)

        # If we do not rely on the L0 regularization, we simply utilize the MSE loss
        if l0 == False:
            loss = torch.nn.functional.mse_loss(next_obs, pred_next_obs)
            total_loss = loss
        # If we on the L0 regularization, we add the L0 penalty to the MSE loss
        else:
            loss = torch.nn.functional.mse_loss(next_obs, pred_next_obs)
            reg = -(model.fc.regularization())
            total_loss = loss + reg

        # compute the loss gradients wrt the model parameters
        total_loss.backward()

        # update the parameters
        optimizer.step()
\end{lstlisting}

\begin{lstlisting}[language=Python, caption=Reward model training loop., label={lst:rew_model_training}]
def train_reward_model(model, optimizer, train_loader, batch_size,                                                num_training_iterations, l0=False):

    for _ in range(num_training_iterations):

        # random sample a batch of data
        data = train_loader.sample_batch(batch_size)

        obs = torch.from_numpy(data['obs']).cuda()
        rew = torch.from_numpy(data['rew']).cuda().reshape(-1, 1)
        act = torch.from_numpy(data['act']).cuda()

        optimizer.zero_grad()

        # predict the reward using the parametric model
        pred_rew = model(obs, act)

        # If we do not rely on the L0 regularization, we simply utilize the MSE loss
        if l0 == False:
            loss = torch.nn.functional.mse_loss(rew, pred_rew)
            total_loss = loss
        # If we on the L0 regularization, we add the L0 penalty to the MSE loss
        else:
            loss = torch.nn.functional.mse_loss(rew, pred_rew)
            reg = -(model.fc.regularization())
            total_loss = loss + reg

        # compute the loss gradients wrt the model parameters
        total_loss.backward()

        # update the parameters
        optimizer.step()
\end{lstlisting}

\subsubsection{Results}
\begin{figure}[h!]
     \centering
     \begin{subfigure}[b]{0.49\textwidth}
         \centering
         \includegraphics[width=\textwidth]{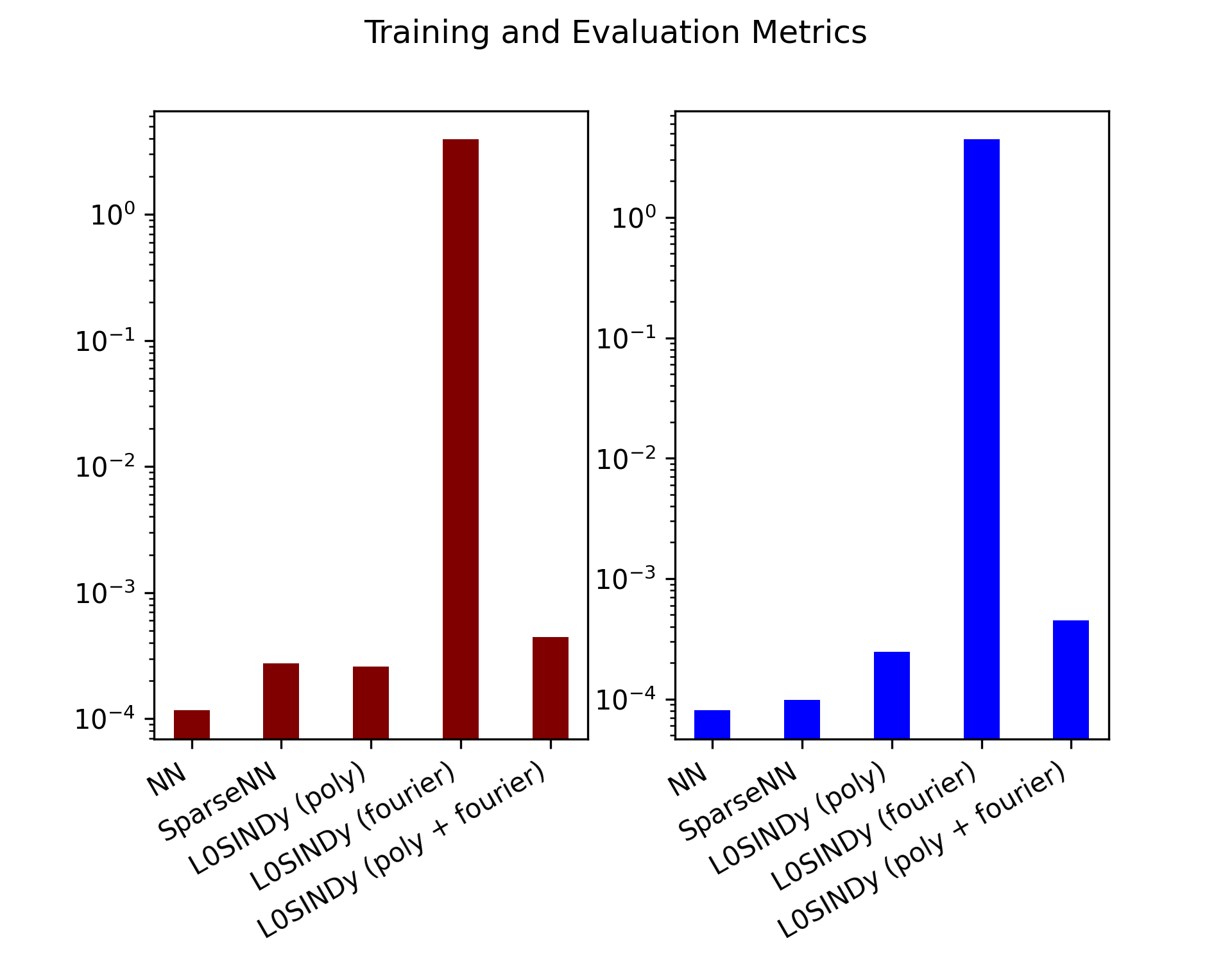}
         \caption{}
         \label{fig:dyn}
     \end{subfigure}
     \begin{subfigure}[b]{0.49\textwidth}
         \centering
         \includegraphics[width=\textwidth]{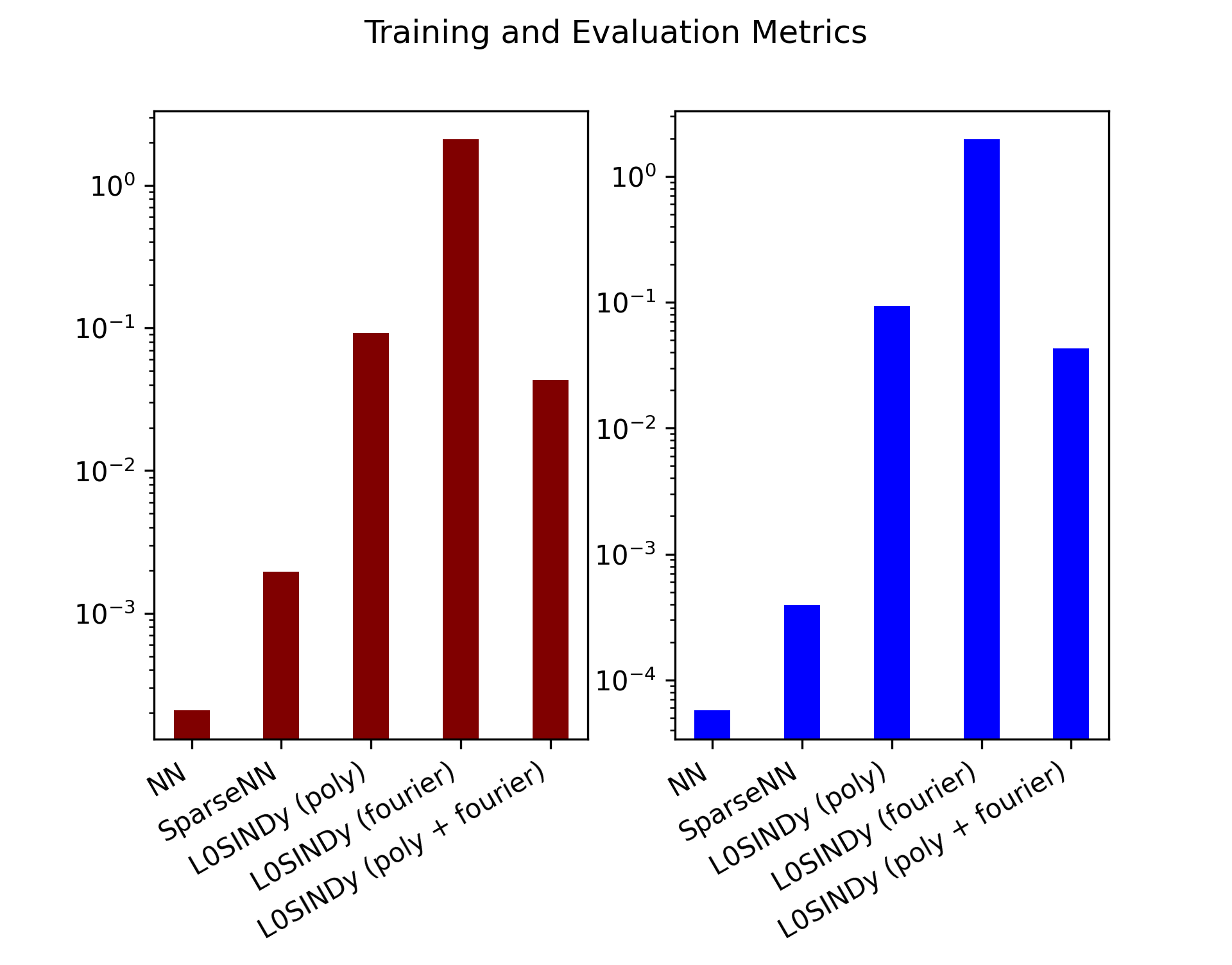}
         \caption{}
         \label{fig:rew}
     \end{subfigure}
        \caption{Prediction errors of the different transition and reward models}
        \label{fig:results_diff_models}
\end{figure}

In Figure \ref{fig:results_diff_models}, we show the accuracy of the different transition and reward models on the training and test sets after 500 training epochs. While the fully-connected neural network seems to be training faster than the sparse models, we need to keep in mind that the L$_0$ regularization makes the minimization of the loss slightly more challenging due to the need for minimizing not only the MSE but also the number of parameters. Therefore, especially for the sparse models, it is worth mentioning that the results can be further improved with longer training time and hyperparameter optimization techniques. 

\subsection{Learning Control Policies}
We utilize the method proposed in \cite{botteghi2024parametric} to learn sparse, and interpretable policies for the pendulum example. We replace the neural network-based policy of the twin-delayed deep deterministic policy gradient algorithm \cite{fujimoto2018addressing} with a polynomial, Fourier, and a polynomial and Fourier policies that are sparsified over training using the L$_0$ regularization (all the details can be found in our paper  \cite{botteghi2024parametric}). It is worth mentioning that our method is independent of the deep reinforcement learning algorithm chosen and other algorithms can be used.

\subsubsection{Results}

\begin{figure}[h!]
    \centering
     \includegraphics[width=0.6\textwidth]{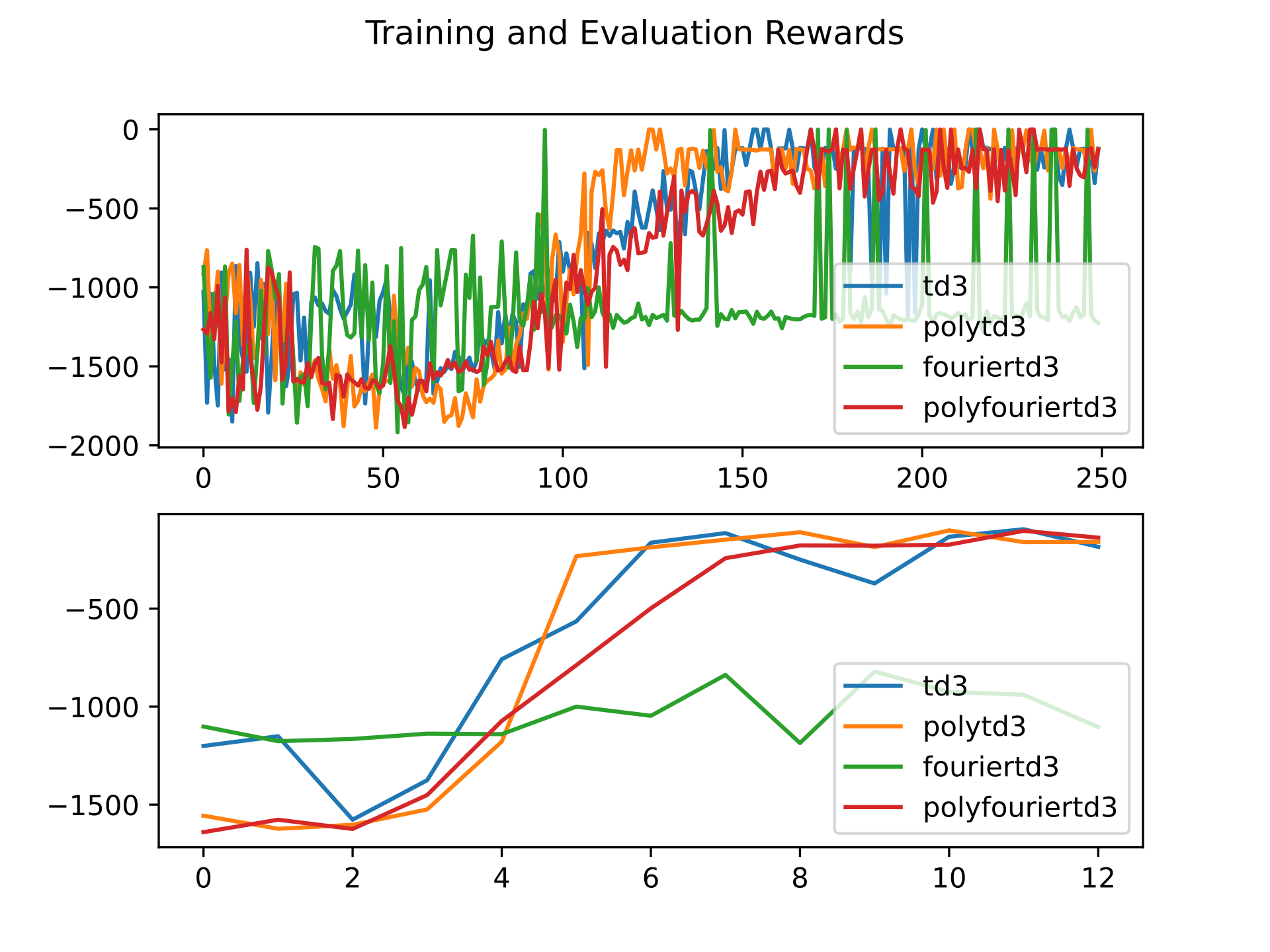}
    \caption{Cumulative reward over training and evaluation.}
    \label{fig:DRL_results}
\end{figure}

In Figure \ref{fig:DRL_results}, we show the training and evaluation rewards collected by the different agents in the simple task of stabilizing the inverted pendulum in its unstable equilibrium. Due to the simplicity of the task, we do not see substantial differences among the agents with the exception of the Fourier agent. This is due to the fact that a policy only composed of sines and cosines is not suitable for a stabilization task in a single point. However, such a policy may be useful for periodic tasks. Due to the limited number of learnable parameters, the agent with polynomial feature library learns to solve the task slightly faster than the fully-connected neural network agent. In addition, the agents relying on the sparse dictionaries allow for deriving a closed-form equation of the learned policies, opening the door to a-posteriori stability and robustness analysis. 

\bibliographystyle{unsrt}
 \begin{spacing}{.9}
 \small{
 \setlength{\bibsep}{2.9pt}
 \bibliography{references}

\begin{thebibliography}{10}

\bibitem{louizos2018learning}
Christos Louizos, Max Welling, and Diederik~P Kingma.
\newblock Learning sparse neural networks through l\_0 regularization.
\newblock In {\em International Conference on Learning Representations}, 2018.

\bibitem{botteghi2024parametric}
Nicol{\`o} Botteghi and Urban Fasel.
\newblock Parametric pde control with deep reinforcement learning and differentiable l0-sparse polynomial policies.
\newblock {\em arXiv preprint arXiv:2403.15267}, 2024.

\bibitem{maddison2016concrete}
Chris~J Maddison, Andriy Mnih, and Yee~Whye Teh.
\newblock The concrete distribution: A continuous relaxation of discrete random variables.
\newblock {\em arXiv preprint arXiv:1611.00712}, 2016.

\bibitem{jang2016categorical}
Eric Jang, Shixiang Gu, and Ben Poole.
\newblock Categorical reparameterization with gumbel-softmax.
\newblock {\em arXiv preprint arXiv:1611.01144}, 2016.

\bibitem{loiseau2020data}
Jean-Christophe Loiseau.
\newblock Data-driven modeling of the chaotic thermal convection in an annular thermosyphon.
\newblock {\em Theoretical and Computational Fluid Dynamics}, 34(4):339--365, 2020.

\bibitem{mitchell1988bayesian}
Toby~J Mitchell and John~J Beauchamp.
\newblock Bayesian variable selection in linear regression.
\newblock {\em Journal of the american statistical association}, 83(404):1023--1032, 1988.

\bibitem{kingma2013auto}
Diederik~P Kingma and Max Welling.
\newblock Auto-encoding variational bayes.
\newblock {\em arXiv preprint arXiv:1312.6114}, 2013.

\bibitem{williams1992simple}
Ronald~J Williams.
\newblock Simple statistical gradient-following algorithms for connectionist reinforcement learning.
\newblock {\em Machine learning}, 8:229--256, 1992.

\bibitem{mnih2014neural}
Andriy Mnih and Karol Gregor.
\newblock Neural variational inference and learning in belief networks.
\newblock In {\em International Conference on Machine Learning}, pages 1791--1799. PMLR, 2014.

\bibitem{mnih2016variational}
Andriy Mnih and Danilo Rezende.
\newblock Variational inference for monte carlo objectives.
\newblock In {\em International Conference on Machine Learning}, pages 2188--2196. PMLR, 2016.

\bibitem{tucker2017rebar}
George Tucker, Andriy Mnih, Chris~J Maddison, John Lawson, and Jascha Sohl-Dickstein.
\newblock Rebar: Low-variance, unbiased gradient estimates for discrete latent variable models.
\newblock {\em Advances in Neural Information Processing Systems}, 30, 2017.

\bibitem{bengio2013estimating}
Yoshua Bengio, Nicholas L{\'e}onard, and Aaron Courville.
\newblock Estimating or propagating gradients through stochastic neurons for conditional computation.
\newblock {\em arXiv preprint arXiv:1308.3432}, 2013.

\bibitem{srinivas2017training}
Suraj Srinivas, Akshayvarun Subramanya, and R~Venkatesh~Babu.
\newblock Training sparse neural networks.
\newblock In {\em Proceedings of the IEEE conference on computer vision and pattern recognition workshops}, pages 138--145, 2017.

\bibitem{gal2017concrete}
Yarin Gal, Jiri Hron, and Alex Kendall.
\newblock Concrete dropout.
\newblock {\em Advances in neural information processing systems}, 30, 2017.

\bibitem{openaigym}
Greg Brockman, Vicki Cheung, Ludwig Pettersson, Jonas Schneider, John Schulman, Jie Tang, and Wojciech Zaremba.
\newblock Openai gym, 2016.

\bibitem{brunton2016discovering}
Steven~L Brunton, Joshua~L Proctor, and J~Nathan Kutz.
\newblock Discovering governing equations from data by sparse identification of nonlinear dynamical systems.
\newblock {\em Proceedings of the national academy of sciences}, 113(15):3932--3937, 2016.

\bibitem{fujimoto2018addressing}
Scott Fujimoto, Herke Hoof, and David Meger.
\newblock Addressing function approximation error in actor-critic methods.
\newblock In {\em International conference on machine learning}, pages 1587--1596. PMLR, 2018.

\end{thebibliography}
 }
 \end{spacing}


\end{document}